\documentclass{sigchi}
\usepackage{pgfplots}
\usepackage{subcaption}
\usetikzlibrary{arrows.meta}
\colorlet{linecol}{black!75}
\usepackage{forest}
\usepackage [english]{babel}
\usepackage [autostyle, english = american]{csquotes}
\MakeOuterQuote{"}
\definecolor{Gray}{gray}{0.9}
\usepackage{color, colortbl}
\usepackage{footnote}
\usepackage{nameref}

\tikzset{my rounded corners/.append style={rounded corners=2pt},}

\conferenceinfo{IUI 2017,}{March 13-16, 2017, Limassol, Cyprus}
\doi{http://dx.doi.org/10.1145/3025171.3025191}

\usepackage{balance}       
\usepackage{graphics}      
\usepackage[T1]{fontenc}   
\usepackage{txfonts}
\usepackage{mathptmx}
\usepackage[pdflang={en-US},pdftex]{hyperref}
\usepackage{color}
\usepackage{booktabs}
\usepackage{textcomp}

\usepackage{microtype}        
\usepackage{ccicons}          

\usepackage{todonotes}

\def\plaintitle{"How May I Help You?": Modeling Twitter Customer Service Conversations Using Fine-Grained Dialogue Acts}

\def\emptyauthor{}
\def\plainkeywords{	Dialogue; Conversation Modeling; Twitter; Customer Service}

\makeatletter
\def\url@leostyle{%
  \@ifundefined{selectfont}{
    \def\UrlFont{\sf}
  }{
    \def\UrlFont{\small\bf\ttfamily}
  }}
\makeatother
\def\pprw{8.5in}
\def\pprh{11in}

\definecolor{linkColor}{RGB}{6,125,233}
\hypersetup{%
  pdftitle={\plaintitle},
  pdfauthor={\emptyauthor},
  pdfkeywords={\plainkeywords},
  pdfdisplaydoctitle=true, 
  bookmarksnumbered,
  pdfstartview={FitH},
  colorlinks,
  citecolor=black,
  filecolor=black,
  linkcolor=black,
  urlcolor=linkColor,
  breaklinks=true,
  hypertexnames=false
}

\begin{document}

\title{\plaintitle}

\numberofauthors{1}
\author{%
\alignauthor
Shereen Oraby$^*$, Pritam Gundecha$^\dag$, Jalal Mahmud$^\dag$, \\ Mansurul Bhuiyan$^\dag$, Rama Akkiraju$^\dag$ \\
       \affaddr{$^*$University of California, Santa Cruz}\\
       \affaddr{$^\dag$IBM Research, Almaden}\\
       \email{soraby@ucsc.edu, \{psgundec $|$ jumahmud $|$ akkiraju\}@us.ibm.com, mansurul.bhuiyan@ibm.com} \\
}

\maketitle

\begin{abstract}
Given the increasing popularity of customer service dialogue on Twitter, analysis of conversation data is essential to understand trends in customer and agent behavior for the purpose of automating customer service interactions. In this work, we develop a novel taxonomy of fine-grained "dialogue acts" frequently observed in customer service, showcasing acts that are more suited to the domain than the more generic existing taxonomies. Using a sequential SVM-HMM model, we model conversation flow, predicting the dialogue act of a given turn in real-time. We characterize differences between customer and agent behavior in Twitter customer service conversations, and investigate the effect of testing our system on different customer service industries. Finally, we use a data-driven approach to predict important conversation outcomes: customer satisfaction, customer frustration, and overall problem resolution. We show that the {\it type} and {\it location} of certain dialogue acts in a conversation have a significant effect on the probability of desirable and undesirable outcomes, and present actionable rules based on our findings. The patterns and rules we derive can be used as guidelines for outcome-driven automated customer service platforms. 
\end{abstract}

\category{H.5.2.}{{\bf User Interfaces}}{Natural Language}


\keywords{\plainkeywords}

\section{Introduction}
The need for real-time, efficient, and reliable customer service has grown in recent years. Twitter has emerged as a popular medium for customer service dialogue, allowing customers to make inquiries and receive instant live support in the public domain. In order to provide useful information to customers, agents must first understand the requirements of the conversation, and offer customers the appropriate feedback. While this may be feasible at the level of a single conversation for a human agent, automatic analysis of conversations is essential for data-driven approaches towards the design of automated customer support agents and systems. 

\begin{table*}[h]
\centering
\small
\begin{tabular}{p{0.7cm}|p{1.2cm}|p{10cm}|p{3.5cm}}
\bf Turn\# & \bf Speaker & \bf Tweet & \bf Relevant Dialogue Acts \\
\hline
\rowcolor{Gray}
1 & \bf Customer & Love my new SMASHED Amiibo box for Mega Man! Priority mail here, guys! /s $<$link$>$ & Complaint, Negative Expressive Statement, Sarcasm \\ \hline
2 & \bf Agent & That's disappointing! Truly sorry. Is the actual $<$item$>$ damaged? & Request Info, Yes-No Question, Apology  \\ \hline
\rowcolor{Gray}
3 & \bf Customer & No, however, I am a collector and I keep them in their boxes. & Negative Answer, Informative Statement  \\ \hline
4 & \bf Agent & I understand. Would you like the item to be exchanged? & Acknowledgement, Yes-No Question, Offer  \\ \hline
\rowcolor{Gray}
5 & \bf Customer &Support If possible, yes. I've never made a return, though. & Affirmative Answer, Informative Statement  \\ \hline
6 & \bf Agent & How did you purchase the item? Did your local store assisted with this order? & Open Question, Request Info, Yes-No Question  \\ \hline
\rowcolor{Gray}
7 & \bf Customer &Support Online. As far as I know this came straight to my house through the website. & Informative Statement  \\ \hline
8 & \bf Agent & If the same item is available at your local store, you may exchange it there. If not, you may call $<$number$>$. & Informative Statement, Suggestion  \\ \hline
\rowcolor{Gray}
9 & \bf Customer &Support Thanks, I'll try it later. & Thanks  \\ \hline
\end{tabular}
\caption{Example Twitter Customer Service Conversation}
\label{sample-conversation}
\end{table*}

Analyzing the dialogic structure of a conversation in terms of the "dialogue acts" used, such as statements or questions, can give important meta-information about conversation flow and content, and can be used as a first step to developing automated agents. Traditional dialogue act taxonomies used to label turns in a conversation are very generic, in order to allow for broad coverage of the majority of dialogue acts possible in a conversation \cite{Core1997,Jurafsky1997,Stolcke2000}. However, for the purpose of understanding and analyzing customer service conversations, generic taxonomies fall short. Table \ref{sample-conversation} shows a sample customer service conversation between a human agent and customer on Twitter, where the customer and agent take alternating "turns" to discuss the problem. As shown from the dialogue acts used at each turn, simply knowing that a turn is a {\it Statement} or {\it Request}, as is possible with generic taxonomies, is not enough information to allow for automated handling or response to a problem. We need more fine-grained dialogue acts, such as {\it Informative Statement}, {\it Complaint}, or {\it Request for Information} to capture the speaker's intent, and act accordingly. Likewise, turns often include multiple overlapping dialogue acts, such that a multi-label approach to classification is often more informative than a single-label approach.

Dialogue act prediction can be used to guide automatic response generation, and to develop diagnostic tools for the fine-tuning of automatic agents. For example, in Table~\ref{sample-conversation}, the customer's first turn (Turn 1) is categorized as a {\it Complaint, Negative Expressive Statement}, and {\it Sarcasm}, and the agent's response (Turn 2) is tagged as a {\it Request for Information, Yes-No Question}, and {\it Apology}. Prediction of these dialogue acts in a real-time setting can be leveraged to generate appropriate automated agent responses to similar situations. 

Additionally, important patterns can emerge from analysis of the fine-grained acts in a dialogue in a post-prediction setting. For example, if an agent does not follow-up with certain actions in response to a customer's {\it question} dialogue act, this could be found to be a violation of a best practice pattern. By analyzing large numbers of dialogue act sequences correlated with specific outcomes, various rules can be derived, i.e. \textit{"Continuing to request information late in a conversation often leads to customer dissatisfaction."} This can then be codified into a best practice pattern rules for automated systems, such as \textit{"A request for information act should be issued early in a conversation, followed by an answer, informative statement, or apology towards the end of the conversation."} 

In this work, we are motivated to predict the dialogue acts in conversations with the intent of identifying problem spots that can be addressed in real-time, and to allow for post-conversation analysis to derive rules about conversation outcomes indicating successful/unsuccessful interactions, namely, customer satisfaction, customer frustration, and problem resolution. We focus on analysis of the dialogue acts used in customer service conversations as a first step to fully automating the interaction. We address various different challenges: dialogue act annotated data is not available for customer service on Twitter, the task of dialogue act annotation is subjective, existing taxonomies do not capture the fine-grained information we believe is valuable to our task, and tweets, although concise in nature, often consist of overlapping dialogue acts to characterize their full intent. The novelty of our work comes from the development of our fine-grained dialogue act taxonomy and multi-label approach for act prediction, as well as our analysis of the customer service domain on Twitter. Our goal is to offer useful analytics to improve outcome-oriented conversational systems.

We first expand upon previous work and generic dialogue act taxonomies, developing a fine-grained set of dialogue acts for customer service, and conducting a systematic user study to identify these acts in a dataset of 800 conversations from four Twitter customer service accounts (i.e. four different companies in the telecommunication, electronics, and insurance industries). We then aim to understand the conversation flow between customers and agents using our taxonomy, so we develop a real-time sequential SVM-HMM model to predict our fine-grained dialogue acts while a conversation is in progress, using a novel multi-label scheme to classify each turn. Finally, using our dialogue act predictions, we classify conversations based on the outcomes of customer satisfaction, frustration, and overall problem resolution, then provide actionable guidelines for the development of automated customer service systems and intelligent agents aimed at desired customer outcomes \cite{allen2001architecture,sun2016intelligent}.

We begin with a discussion of related work, followed by an overview of our methodology. Next, we describe our conversation modeling framework, and explain our outcome analysis experiments, to show how we derive useful patterns for designing automated customer service agents. Finally, we present conclusions and directions for future work. 

\section{Related Work} \label{related-work}
Developing computational speech and dialogue act models has long been a topic of interest \cite{Austin,Morelli,Sacks,Searle}, with researchers from many different backgrounds studying human conversations and developing theories around conversational analysis and interpretation on intent. Modern intelligent conversational~\cite{allen2001architecture,sun2016intelligent} and dialogue systems draw principles from many disciplines, including philosophy, linguistics, computer science, and sociology. In this section, we describe relevant previous work on speech and dialogue act modeling, general conversation modeling on Twitter, and speech and dialogue act modeling of customer service in other data sources.

Previous work has explored speech act modeling in different domains (as a predecessor to dialogue act modeling). Zhang et al. present work on recognition of speech acts on Twitter, following up with a study on scalable speech act recognition given the difficulty of obtaining labeled training data \cite{Zhang2011}. They use a simple taxonomy of four main speech acts ({\it Statement, Question, Suggestion, Comment}, and a {\it Miscellaneous} category). More recently, Vosoughi et al. develop~\cite{vosoughi2016tweet} a speech act classifier for Twitter, using a modification of the taxonomy defined by Searle in 1975, including six acts they observe to commonly occur on Twitter: {\it Assertion, Recommendation Expression, Question, Request}, again plus a {\it Miscellaneous} category. They describe good features for speech act classification and the application of such a system to detect stories on social media \cite{Vosoughi2013}. In this work, we are interested in the dialogic characteristics of Twitter conversations, rather than speech acts in stand-alone tweets.

Different dialogue act taxonomies have been developed to characterize conversational acts. Core and Allen present the Dialogue Act Marking in Several Layers (DAMSL), a standard for discourse annotation that was developed in 1997 \cite{Core1997}. The taxonomy contains a total of 220 tags, divided into four main categories: communicative status, information level, forward-looking function, and backward-looking function. Jurafsky, Shriberg, and Biasca develop a less fine-grained taxonomy of 42 tags based on DAMSL \cite{Jurafsky1997}. Stolcke et al. employ a similar set for general conversation \cite{Stolcke2000}, citing that "content- and task-related distinctions will always play an important role in effective DA [Dialogue Act] labeling." Many researchers have tackled the task of developing different speech and dialogue act taxonomies and coding schemes \cite{Bunt2010, Kluwer2010, Schiffrin2005, Tur2006}. For the purposes of our own research, we require a set of dialogue acts that is more closely representative of customer service domain interactions - thus we expand upon previously defined taxonomies and develop a more fine-grained set.

Modeling general conversation on Twitter has also been a topic of interest in previous work. Honeycutt and Herring study conversation and collaboration on Twitter using individual tweets containing "@" mentions \cite{Honey}. Ritter et al. explore unsupervised modeling of Twitter conversations, using clustering methods on a corpus of 1.3 million Twitter conversations to define a model of transitional flow between in a general Twitter dialogue \cite{Ritter}. While these approaches are relevant to understanding the nature of interactions on Twitter, we find that the customer service domain presents its own interesting characteristics that are worth exploring further.

The most related previous work has explored speech and dialogue act modeling in customer service, however, no previous work has focused on Twitter as a data source. In 2005, Ivanovic uses an abridged set of 12 course-grained dialogue acts (detailed in the Taxonomy section) to describe interactions between customers and agents in instant messaging chats \cite{Ivanovic2005,ivanovic2008automatic}, leading to a proposal on response suggestion using the proposed dialogue acts \cite{Ivanovic2006}. Follow-up work using the taxonomy selected by Ivanovic comes from Kim et al., where they focus on classifying dialogue acts in both one-on-one and multi-party live instant messaging chats \cite{Kim2012, Kim2012a}. These works are similar to ours in the nature of the problem addressed, but we use a much more fine-grained taxonomy to define the interactions possible in the customer service domain, and focus on Twitter conversations, which are unique in their brevity and the nature of the public interactions.

The most similar work to our own is that of Herzig et al. on classifying emotions in customer support dialogues on Twitter \cite{herzig}. They explore how agent responses should be tailored to the detected emotional response in customers, in order to improve the quality of service agents can provide. Rather than focusing on emotional response, we seek to model the dialogic structure and intents of the speakers using dialogue acts, with emotion included as features in our model, to characterize the emotional intent within each act.

\section{Methodology}
\label{methodology}
\begin{figure}[h!]
  \centering
    \includegraphics[width=0.25\textwidth]{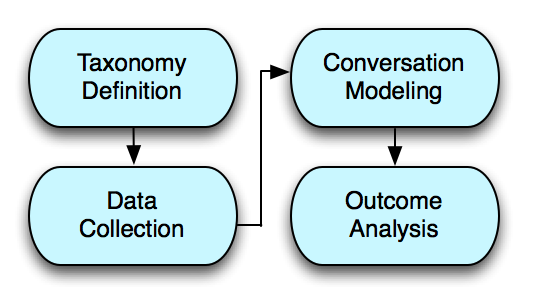}
\caption{Methodology Pipeline}
\label{method-pipeline}
\end{figure}

The underlying goal of this work is to show how a well-defined taxonomy of dialogue acts can be used to summarize semantic information in real-time about the flow of a conversation to derive meaningful insights into the success/failure of the interaction, and then to develop actionable rules to be used in automating customer service interactions. We focus on the customer service domain on Twitter, which has not previously been explored in the context of dialogue act classification. In this new domain, we can provide meaningful recommendations about good communicative practices, based on real data. Our methodology pipeline is shown in Figure \ref{method-pipeline}.

\begin{enumerate}
  \setlength\itemsep{-0.1em}
\item {\bf Taxonomy Definition and Data Collection:} We expand on previous work by defining a taxonomy of fine-grained dialogue acts suited to the customer service domain, and use this taxonomy to gather annotations for customer service conversations on Twitter.
\item {\bf Conversation Modeling:} We develop an SVM-HMM model to identify the different dialogue acts in a customer service conversation, in a real-time setting, and using a novel multi-label approach to capture different dialogic intents contained in a single turn. We compare the performance of the model under different settings to better understand differences between customer and agent behavior.
\item {\bf Conversation Outcome Analysis:} We use our model to provide actionable recommendations for the development of automated customer service systems, answering questions such as, "What is the correlation between conversation flow in terms of the dialogue acts used, and overall customer satisfaction, frustration, and problem resolution?", and "What rules can we include in automated systems to promote successful interactions with customers?"
\end{enumerate}

\section{Taxonomy Definition}
\label{taxonomy}

\begin{figure*}[t!]
\small
\centering
    \begin{forest}
      for tree={
        line width=1pt,
        if={level()<2}{
          my rounded corners,
          draw=linecol,
        }{},
        fit=rectangle,
        edge={color=linecol, >={Triangle[]}, },
        if level=0{%
          l sep+=0.2cm,
          for descendants={%
            calign=first,
          },
          align=center,
          parent anchor=south,
        }{%
          if level=1{%
            parent anchor=south west,
            child anchor=north,
            tier=three ways,
            align=center,
            for descendants={%
              child anchor=west,
              parent anchor=west,
              align=left,
              anchor=west,
              xshift=-15pt,
              edge path={
                \noexpand\path[\forestoption{edge}]
                (!to tier=three ways.parent anchor) |-
                (.child anchor)\forestoption{edge label};
              },
            },
          }{}%
        },
      }
      [Proposed Fine-Grained Dialogue Act Taxonomy
        [Greeting 
          [Opening \\ Closing
          ]
        ]
        [Statement 
            [Informative \\ Expressive Positive \\ Expressive Negative \\ Complaint \\ Offer Help \\ Suggest Action \\ Promise \\ Sarcasm \\ Other            ]
        ]
        [Request 
            [Request Help \\ Request Info \\ Other          ]
        ]
        [Question 
            [Yes-No Question \\ Wh- Question \\ Open Question 
              ]
        ]
        [Answer 
            [Yes-Answer \\ No-Answer \\ Response-Ack \\ Other
                 ]
                 ]
        [Social Act 
            [Thanks \\ Apology \\ Downplayer
           ] ]
      ]
    \end{forest}
\caption{Proposed Fine-Grained Dialogue Act Taxonomy for Customer Service}
\label{fig:new-taxonomy}
\end{figure*}
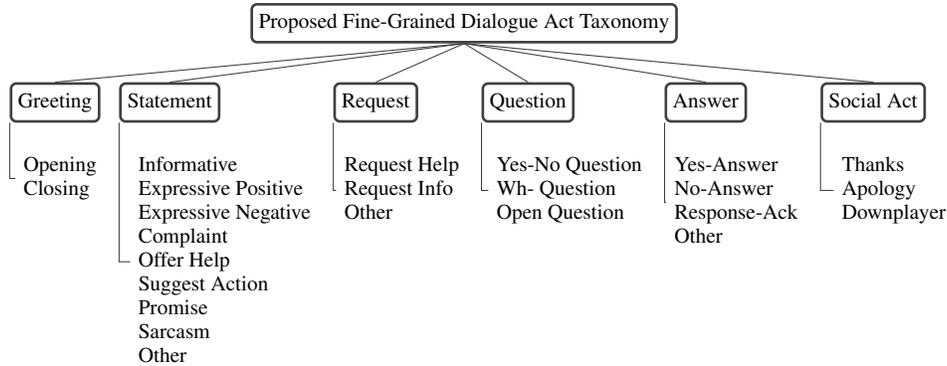

As described in the related work, the taxonomy of 12 acts to classify dialogue acts in an instant-messaging scenario, developed by Ivanovic in 2005, has been used by previous work when approaching the task of dialogue act classification for customer service \cite{Ivanovic2005, Ivanovic2006, ivanovic2008automatic, Kim2012, Kim2012a}.  The dataset used consisted of eight conversations from chat logs in the MSN Shopping Service (around 550 turns spanning around 4,500 words) \cite{ivanovic2008automatic}. The conversations were gathered by asking five volunteers to use the platform to inquire for help regarding various hypothetical situations (i.e. buying an item for someone) \cite{ivanovic2008automatic}. The process of selection of tags to develop the taxonomy, beginning with the 42 tags from the DAMSL set \cite{Core1997}, involved removing tags inappropriate for written text, and collapsing sets of tags into a more coarse-grained label \cite{Ivanovic2005}. The final taxonomy consists of the following 12 dialogue acts (sorted by frequency in the dataset): {\it Statement (36\%), Thanking (14.7\%), Yes-No Question (13.9\%), Response-Acknowledgement (7.2\%), Request (5.9\%), Open-Question (5.3\%), Yes-Answer (5.1\%), Conventional-Closing (2.9\%), No-Answer (2.5\%), Conventional-Opening (2.3\%), Expressive (2.3\%)} and {\it Downplayer (1.9\%)}.


For the purposes of our own research, focused on customer service on Twitter, we found that the course-grained nature of the taxonomy presented a natural shortcoming in terms of what information could be learned by performing classification at this level. We observe that while having a smaller set of dialogue acts may be helpful for achieving good agreement between annotators (Ivanovic cites kappas of 0.87 between the three expert annotators using this tag set on his data \cite{Ivanovic2005}), it is unable to offer deeper semantic insight into the specific intent behind each act for many of the categories. For example, the {\it Statement} act, which comprises the largest percentage (36\% of turns), is an extremely broad category that fails to provide useful information from an analytical perspective. Likewise, the {\it Request} category also does not specify any intent behind the act, and leaves much room for improvement.

For this reason, and motivated by previous work seeking to develop dialogue act taxonomies appropriate for different domains \cite{ivanovic2008automatic,Kim2012}, we convert the list of dialogue acts presented by the literature into a hierarchical taxonomy, shown in Figure~\ref{fig:new-taxonomy}. 

We first organize the taxonomy into six high-level dialogue acts: {\it Greeting, Statement, Request, Question, Answer,} and {\it Social Act}. Then, we update the taxonomy using two main steps: restructuring and adding additional fine-grained acts.

\begin{itemize}
  \setlength\itemsep{0.5em}
\item We restructure 10 of the acts into our higher-level categories:  {\it Conventional Opening} and {\it Conventional Closing} into {\it Greeting}; {\it Expressive} into {\it Statement} (further dividing it into {\it Expressive Positive} and {\it Expressive Negative}); {\it Yes-No Question} and {\it Open Question} into {\it Question}; {\it Yes-Answer}, {\it No Answer}, and {\it Response-Ack} into {\it Answer}; and {\it Thanking} and {\it Downplayer} into {\it Social Act}.
\item Next, we add fine-grained acts to the two very broad categories of {\it Statement} and {\it Request}. We add {\it Giving Information, Complaint, Offer Help, Suggest Action, Promise, Sarcasm,} and {\it Other} categories to {\it Statement}, and {\it Request Help, Request Info,} and {\it Other} categories to {\it Request}.
\end{itemize}

We base our changes upon the taxonomy used by Ivanovic and Kim et al. in their work on instant messaging chat dialogues \cite{ivanovic2008automatic,Kim2012}, but also on general dialogue acts observed in the customer service domain, including {\it complaints} and {\it suggestions}. Our taxonomy does not make any specific restrictions on which party in the dialogue may perform each act, but we do observe that some acts are far more frequent (and sometimes non-existent) in usage, depending on whether the customer or agent is the speaker (for example, the {\it Statement Complaint} category never shows up in Agent turns).

In order to account for gaps in available act selections for annotators, we include an {\it Other} act in the broadest categories. While our taxonomy fills in many gaps from previous work in our domain, we do not claim to have handled coverage of all possible acts in this domain. Our taxonomy allows us to more closely specify the intent and motivation behind each turn, and ultimately how to address different situations. 

\section{Data Collection}
\label{data}
Given our taxonomy of fine-grained dialogue acts that expands upon previous work, we set out to gather annotations for Twitter customer service conversations.

For our data collection phase, we begin with conversations from the Twitter customer service pages of four different companies,\footnote{We keep the names of the companies anonymous, replacing them with placeholders in our annotation tasks.} from the electronics, telecommunications, and insurance industries.\footnote{The conversations were provided to us by Herzig et al. from the same pool used in their work \cite{herzig}.} We perform several forms of pre-processing to the conversations. We filter out conversations if they contain more than one customer or agent speaker, do not have alternating customer/agent speaking turns (single turn per speaker), have less than 5 or more than 10 turns,{\footnote{{The lower bound was set to allow for at least 2 turns per speaker, and the upper-bound was selected after finding that 93\% of the conversations had 10 or fewer turns}}} have less than 70 words in total, and if any turn in the conversation ends in an ellipses followed by a link (indicating that the turn has been cut off due to length, and spans another tweet). Additionally, we remove any references to the company names (substituting with "{\tt Agent}"), any references to customer usernames (substituting with "{\tt Customer}"), and replacing and links or image references with {\tt $<$link$>$} and {\tt $<$img$>$} tokens. 

Using these filters as pre-processing methods, we end up with a set of 800 conversations, spanning 5,327 turns. We conduct our annotation study on Amazon Mechanical Turk,\footnote{\url{http://www.mturk.com}} presenting Turkers with Human Intelligence Tasks (henceforth, HITs) consisting of a single conversation between a customer and an agent. In each HIT, we present Turkers with a definition of each dialogue act, as well as a sample annotated dialogue for reference. For each turn in the conversation, we allow Turkers to select as many labels from our taxonomy as required to {\it fully characterize the intent of the turn}. Additionally, annotators are asked three questions at the end of each conversation HIT, to which they could respond that they agreed, disagreed, or could not tell: 
\begin{itemize}
  \setlength\itemsep{0.5em}
\item {\it At any point in the conversation, does the customer seem frustrated?} 
\item {\it By the end of the conversation, does the customer seem satisfied?}
\item {\it By the end of the conversation, was the problem resolved (or will the parties continue the conversation)?}
\end{itemize}

We ask 5 Turkers to annotate each conversation HIT, and pay \$0.20 per HIT.  We find the list of "majority dialogue acts" for each tweet by finding any acts that have received majority-vote labels (at least 3 out of 5 judgements).

It is important to note at this point that we make an important choice as to how we will handle dialogue act tagging for each turn. We note that each turn may contain more than one dialogue act vital to carry its full meaning. Thus, we choose {\bf not} to carry out a specific segmentation task on our tweets, contrary to previous work \cite{manuvinakurike,Zarisheva2015}, opting to characterize each tweet as a single unit composed of different, often overlapping, dialogue acts. Table \ref{tweet-samples} shows examples of tweets that receive majority vote on more than one label, where the act boundaries are overlapping and not necessarily distinguishable.

\begin{table}[h!]
\centering
\small
\caption{Sample Tweets with Overlapping Dialogue Acts}
\label{tweet-samples}
\begin{tabular}{p{0.1cm}|p{7.8cm}}
\hline
1 & {\it @Customer That's not what we like to hear. What's causing u to feel this way? How can we turn this around for u? We're here to help. } \\
 & {\bf Statement Offer, Request for Info, Question Open}\\
 \hline  \hline
2 & {\it "Thanks @Agent for screwing me over again. Once I'm done figuring out the problems you caused me, I'll be taking my services elsewhere."} \\ 
& {\bf Statement Informative, Statement Complaint, Statement Sarcasm} \\
\hline
\end{tabular}
\end{table}

It is clear that the lines differentiating these acts are not very well defined, and that segmentation would not necessarily aid in clearly separating out each intent. For these reasons, and due to the overall brevity of tweets in general, we choose to avoid the overhead of requiring annotators to provide segment boundaries, and instead ask for all appropriate dialogue acts. 

\begin{figure}[b!]
  \centering
    \includegraphics[width=0.47\textwidth]{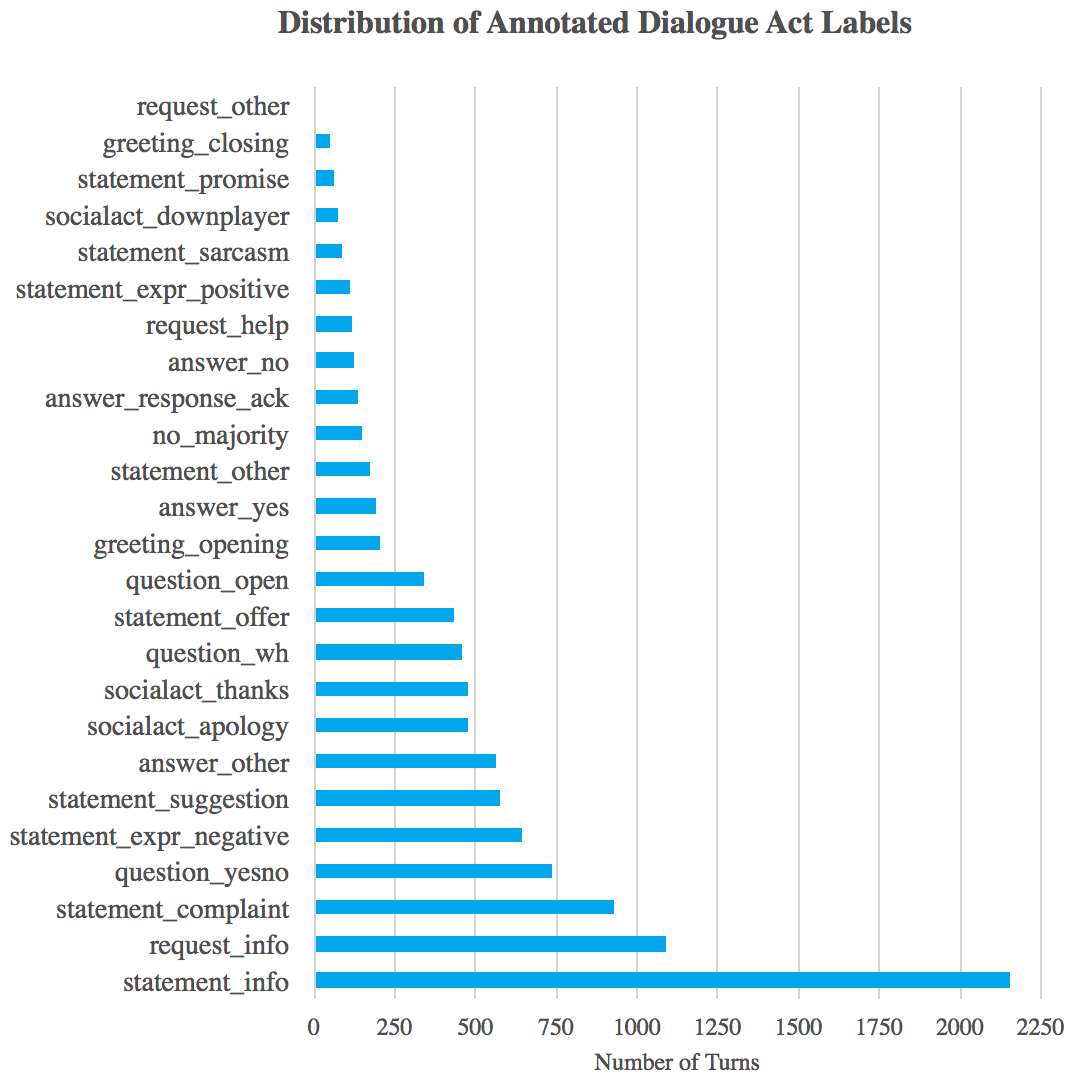}
\caption{Distribution of Annotated Dialogue Act Labels}
\label{distribution-plot}
\end{figure}

{ \begin{table}[t]
\centering
\small
\caption{{ Dialogue Act Agreement in Fleiss-$\kappa$ Bins (from Landis and Koch, 1977) }}
\label{label-bins-kappa}
\begin{tabular}{p{2cm}|p{5.5cm}}
{\bf Agreement} & {\bf Dialogue Acts} \\
\hline
Slight \:\:\:\:\:\:\:\:\:\:\: (0.01-0.20) & Statement Other, Answer Response Ack, Request Other\\ \hline
Fair \:\:\:\:\:\:\:\:\:\:\:\:\:\:\:\:\:\:\:\:\:\: (0.21-0.40) & Statement Sarcasm, Answer Other, Statement Promise, Greeting Closing, Question Open, Statement Expressive Pos. \\ \hline
Moderate \:\:\:\:\:\:\:\:\:\:\:(0.41-0.60) & Statement Complaint, Question Wh-, Social Act Downplayer, Statement Offer, Request Info, Statement Info, Request Help, Statement Expressive Neg.\\ \hline
Substantial \:\:\:\:\:\:\:\:\:\:\: (0.61-0.80) & Greeting Opening, Question Yes-No, Answer Yes, Answer No, Statement Suggestion \\ \hline
Almost Perfect (0.81-1.00) & Social Act Apology, Social Act Thanks\\
\end{tabular}
\end{table}}

\begin{table*}[t!]
\centering
\small
\caption{Detailed Distribution of Top 12 Fine-Grained Dialogue Acts Derived From Annotations}
\label{annotation-distribution}
\begin{tabular}{p{3.8cm}|p{6.5cm}|p{2cm}|p{2cm}}
\bf Tag & \bf Example & \bf \% of Turns \:\:\: (5,327) & \bf \% of Annot. (10,343) \\ \hline
Statement Informative & The signal came back last night [...] & 40.3 & 20.8 \\
Request Information & Can you send us [...]? & 20.4 & 10.5 \\
Statement Complaint & Staff didn't honor online info, was dismissive [...] & 17.3 & 8.9 \\
Question Yes-No & Have you tried for availability at [...] & 13.7 & 7.0 \\
Statement Expressive Neg. & I don't trust places that do bad installations [...] & 12.0 & 6.2 \\
Statement Suggestion & Let's try clearing the cache $<$link$>$ [...] & 10.7 & 5.5 \\
Answer (Other) & Depends on the responder [...] & 10.5 & 5.4 \\
Social Act Apology & I'm sorry for the trouble [...] & 8.8 & 4.5 \\
Social Act Thanks & Thanks for the help [...] & 8.8 & 4.5 \\
Question Wh- & Why was that? & 8.5 & 4.4 \\
Statement Offer & We can always double check the account [...] & 8.1 & 4.1 \\
Question Open & How come I can't get a [...] quote online? & 6.4 & 3.3 \\  \hline
(All Other Acts) & &  2.7 & 14.3 \\
\end{tabular}
\end{table*}

\subsection{Annotation Results}
Figure \ref{distribution-plot} shows the distribution of the number of times each dialogue act in our taxonomy is selected a majority act by the annotators (recall that each turn is annotated by 5 annotators). From the distribution, we see that the largest class is {\it Statement Info} which is part of the majority vote list for 2,152 of the 5,327 total turns, followed by {\it Request Info}, which appears in 1,088 of the total turns. Although {\it Statement Informative} comprises the largest set of majority labels in the data (as did {\it Statement} in Ivanovic's distribution), we do observe that other fine-grained categories of {\it Statement} occur in the most frequent labels as well, including {\it Statement Complaint}, {\it Statement Expressive Negative}, and {\it Statement Suggestion} -- giving more useful information as to what form of statement is most frequently occurring. We find that 147 tweets receive no majority label (i.e. no single act received 3 or more votes out of 5). At the tail of the distribution, we see less frequent acts, such as {\it Statement Sarcasm, Social Act Downplayer, Statement Promise, Greeting Closing, and Request Other}. It is also interesting to note that both opening and closing greetings occur infrequently in the data -- which is understandable given the nature of Twitter conversation, where formal greeting is not generally required.

Table \ref{annotation-distribution} shows a more detailed summary of the distribution of our top 12 dialogue acts according to the annotation experiments, as presented by Ivanovic \cite{Ivanovic2005}. Since each turn has an overlapping set of labels, the column {\bf \% of Turns (5,327)} represents what fraction of the total 5,327 turns contain that dialogue act label (these values do not sum to 1, since there is overlap). To give a better sense of the percentage appearance of each dialogue act class in terms of the total number of annotated labels given, we also present column {\bf \% of Annotations (10,343)} (these values are percentages). 

We measure agreement in our annotations using a few different techniques. Since each item in our annotation experiments allows for multiple labels, we first design an agreement measure that accounts for how frequently each annotator selects the acts that agree with the majority-selected labels for the turns they annotated. { To calculate this for each annotator, we find the number of majority-selected acts for each conversation they annotated (call this {\tt MAJ}), and the number of subset those acts that they selected (call this {\tt SUBS}), and find the ratio ({\tt SUBS/MAJ}). We use this ratio to systematically fine-tune our set of annotators by running our annotation in four batches, restricting our pool of annotators to those that have above a 0.60 ratio of agreement with the majority from the previous batch, as a sort of quality assurance test. }We also measure Fleiss' Kappa \cite{fleiss} agreement between annotators in two ways: first by normalizing our annotation results into binary-valued items indicating annotators' votes for each label contain within each turn. We find an average Fleiss-$\kappa=0.528$ for the full dataset, including all turn-and-label items, representing moderate agreement on the 24-label problem. 

{ We also calculate the Fleiss-$\kappa$ values for each label, and use the categories defined by Landis and Koch to bin our speech acts based on agreement \cite{Landis77}. As shown in Table \ref{label-bins-kappa}, we find that the per-label agreement varies from "almost perfect" agreement of $\kappa=0.871$ for lexically defined categories such as {\it Apology} and {\it Thanks}, with only slight agreement of $\kappa=0.01-0.2$ for less clearly-defined categories, such as {\it Statement (Other), Answer Response Acknowledgement} and {\it Request (Other)}. For the conversation-level questions, we calculate the agreement across the "Agree" label for all annotators, finding an average Fleiss-$\kappa=0.595$, with question-level results of $\kappa=0.624$ for customer satisfaction, $\kappa=0.512$ for problem resolution, and $\kappa=0.384$ for customer frustration. These results suggest room for improvement for further development of the taxonomy, to address problem areas for annotators and remedy areas of lower agreement.}

\subsection{Motivation for Multi-Label Classification}
\label{overlapping-labels}

We test our hypothesis that tweet turns are often characterized by more than one distinct dialogue act label by measuring the percentage overlap between frequent pairs of labels. Of the 5,327 turns annotated, across the 800 conversations, we find that 3,593 of those turns (67.4\%) contained more than one majority-act label. Table \ref{pair-plot} shows the distribution percentage of the most frequent pairs. 

\begin{table}[h!]
\centering
\small
\caption{Distribution of the 10 Most Frequent Dialogue Act Pairs for Turns with More Than 1 Label (3,593)}
\label{pair-plot}
\begin{tabular}{p{5.5cm}|p{1.5cm}}
\bf Dialogue Act Pair & \bf \% of Turns \\ \hline
(statement\_info, answer\_other) & 13.74 \\
(statement\_expr\_neg, statement\_complaint) & 12.71 \\
(statement\_info, statement\_complaint) & 12.10 \\
(request\_info, question\_yesno) & 9.18 \\
(request\_info, question\_wh) & 8.26 \\
(statement\_offer, request\_info) & 5.17 \\
(statement\_info, statement\_expr\_neg) & 4.81 \\
(request\_info, socialact\_apology) & 4.75 \\
(statement\_info, statement\_suggestion) & 4.39 \\
\end{tabular}
\end{table}

For example, we observe that answering with informative statements is the most frequent pair, followed by complaints coupled with negative sentiment or informative statements. We also observe that requests are usually formed as questions, but also co-occur frequently with apologies. This experiment validates our intuition that the majority of turns do contain more than a single label, and motivates our use of a multi-label classification method for characterizing each turn in the conversation modeling experiments we present in the next section.


\section{Conversation Modeling}
\label{experiments}
In this section, we describe the setup and results of our conversational modeling experiments on the data we collected using our fine-grained taxonomy of customer service dialogue acts. We begin with an overview of the features and classes used, followed by our experimental setup and results for each experiment performed.

\subsection{Features}
\label{sec-features}
The following list describes the set of features used for our dialogue act classification tasks:

\begin{itemize}  \setlength\itemsep{0.5em}
\item {\bf Word/Punctuation}: binary bag-of-word unigrams, binary existence of a question mark, binary existence of an exclamation mark in a turn
\item {\bf Temporal:} response time of a turn (time in seconds elapsed between the posting time of the previous turn and that of the current turn)  
\item {\bf Second-Person Reference:} existence of an explicit second-person reference in the turn (you, your, you're)
\item {\bf Emotion:} count of words in each of the 8 emotion classes from the NRC emotion lexicon \cite{NRC} ({\it anger, anticipation, disgust, fear, joy, negative, positive, sadness, surprise}, and {\it trust})
\item {\bf Dialogue:} lexical indicators in the turn: opening greetings (hi, hello, greetings, etc), closing greetings (bye, goodbye), yes-no questions (turns with questions starting with do, did, can, could, etc), wh- questions (turns with questions starting with who, what, where, etc), thanking (thank*), apology (sorry, apolog*), yes-answer, and no-answer
\end{itemize}

\subsection{Classes}
\label{sec-classes}
Table \ref{classes} shows the division of classes we use for each of our experiments. We select our classes using the distribution of annotations we observe in our data collection phase (see Table \ref{annotation-distribution}), selecting the top 12 classes as candidates. 

\begin{table}[h!]
\centering
\small
\caption{Dialogue Acts Used in Each Set of Experiments}
\label{classes}
\begin{tabular}{p{1.5cm}|p{3.5cm}}
{\bf Name} & {\bf Dialogue Acts} \\
\hline
{\bf 6 Class} & Statement Informative \\
&  Request Information \\
& Statement Complaint \\
& Question Yes-No \\
& Statement Expressive Negative \\
& (All Other Acts)\\ \hline
{\bf 8 Class} & {\bf 6-Class} + \\
& Statement Suggestion \\
& Statement Answer Other \\ \hline
{\bf 10-Class} & {\bf 8-Class} + \\
{\bf (Easy)} & Social Act Apology \\
& Social Act Thanks \\ \hline
{\bf 10-Class} & {\bf 8-Class} + \\
{\bf (Hard)} & Statement Offer \\
& Question Open\\
\hline
\end{tabular}
\end{table}

While iteratively selecting the most frequently-occurring classes helps to ensure that classes with the most data are represented in our experiments, it also introduces the problem of including classes that are very well-defined lexically, and may not require learning for classification, such as {\it Social Act Apology} and {\it Social Act Thanking} in the first 10-Class set. For this reason, we call this set 10-Class (Easy), and also experiment using a 10-Class (Hard) set, where we add in the next two less-defined and more semantically rich labels, such as {\it Statement Offer} and {\it Question Open}. When using each set of classes, a turn is either classified as one of the classes in the set, or it is classified as "other" (i.e. any of the other classes). We discuss our experiments in more detail and comment on performance differences in the experiment section.


\subsection{Experiments}
Following previous work on conversation modeling \cite{herzig}, we use a sequential SVM-HMM (using the $SVM^{HMM}$ toolkit \cite{altun}) for our conversation modeling experiments. We hypothesize that a sequential model is most suited to our dialogic data, and that we will be able to concisely capture conversational attributes such as the order in which dialogue acts often occur (i.e. some {\it Answer} act after {\it Question} a question act, or {\it Apology} acts after {\it Complaints}).

We note that with default settings for a sequence of length $N$, an SVM-HMM model will be able to refine its answers for any turn $x_i$ as information becomes available for turns $x_{i+1...N}$. However, we opt to design our classifier under a real-time setting, where turn-by-turn classification is required {\it without} future knowledge or adaptation of prediction at any given stage. In our setup, turns are predicted in a real-time setting to fairly model conversation available to an intelligent agent in a conversational system. At any point, a turn $x_i$ is predicted using information from turns $x_{1...i}$, and where a prediction is not changed when new information is available. 

We test our hypothesis by comparing our real-time sequential SVM-HMM model to non-sequential baselines from the NLTK \cite{nltk} and Scikit-Learn \cite{scikit-learn} toolkits. We use our selected feature set (described above) to be generic enough to apply to both our sequential and non-sequential models, in order to allow us to fairly compare performance. We shuffle and divide our data into 70\% for training and development (560 conversations, using 10-fold cross-validation for parameter tuning), and hold out 30\% of the data (240 conversations) for test.

Motivated by the prevalent overlap of dialogue acts, we conduct our learning experiments using a multi-label setup. For each of the sets of classes, we conduct binary classification task for each label: for each $N$-class classification task, a turn is labeled as either belonging to the current label, or not (i.e. "other"). In this setup, each turn is assigned a binary value for each label (i.e. for the 6-class experiment, each turn receives a value of 0/1 for each indicating whether the classifier predicts it to be relevant to the each of the 6 labels). Thus, for each $N$-class experiment, we end up with $N$ binary labels, for example, whether the turn is a {\it Statement Informative} or {\it Other}, {\it Request Information} or {\it Other}, etc. 

We aggregate the $N$ binary predictions for each turn, then compare the resultant prediction matrix for all turns to our majority-vote ground-truth labels, where at least 3 out of 5 annotators have selected a label to be true for a given turn. The difficulty of the task increases as the number of classes $N$ increases, as there are more classifications done for each turn (i.e., for the 6-class problem, there are 6 classification tasks per turn, while for the 8-class problem, there are 8, etc). Due to the inherent imbalance of label-distribution in the data (shown in Figure \ref{distribution-plot}), we use weighted F-macro to calculate our final scores for each feature set (which finds the average of the metrics for each label, weighted by the number of true instances for that label) \cite{scikit-learn}. 

\subsubsection{Non-Sequential Baselines vs. Sequential SVM-HMM}
\begin{figure}[h!]
  \centering 
    \includegraphics[width=0.5\textwidth]{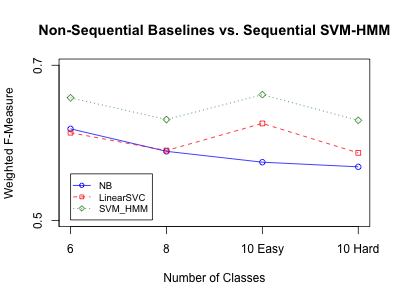}
      \caption{Plot of Non-Sequential Baselines vs. Sequential SVM-HMM Model}
      \label{exp-baselines}
\end{figure}

Our first experiment sets out to compare the use of a non-sequential classification algorithm versus a sequential model for dialogue act classification on our dataset. We experiment with the default Naive Bayes (NB) and Linear SVC algorithms from Scikit-Learn \cite{scikit-learn}, comparing with our sequential SVM-HMM model. We test each classifier on each of our four class sets, reporting weighted F-macro for each experiment. Figure \ref{exp-baselines} shows the results of the experiments. 

From this experiment, we observe that our sequential SVM-HMM outperforms each non-sequential baseline, for each of the four class sets. We select the sequential SVM-HMM model for our preferred model for subsequent experiments. We observe that while performance may be expected to drop as the number of classes increases, we instead get a spike in performance for the 10-Class (Easy) setting. This increase occurs due to the addition of the lexically well-defined classes of {\it Statement Apology} and {\it Statement Thanks}, which are much simpler for our model to predict. Their addition results in a performance boost, comparable to that of the simpler 6-Class problem. When we remove the two well-defined classes and add in the next two broader dialogue act classes of {\it Statement Offer} and {\it Question Open} (as defined by the 10-Class (Hard) set), we observe a drop in performance, and an overall result comparable to our 8-Class problem. This result is still strong, since the number of classes has increased, but the overall performance does not drop. 

We also observe that while NB and LinearSVC have the same performance trend for the smaller number of classes, Linear SVC rapidly improves in performance as the number of classes increases, following the same trend as SVM-HMM. The smallest margin of difference between SVM-HMM and Linear SVC also occurs at the 10-Class (Easy) setting, where the addition of highly-lexical classes makes for a more differentiable set of turns.

\subsubsection{Customer-Only vs. Agent-Only Turns}
\begin{figure}[h!]
  \centering
    \includegraphics[width=0.5\textwidth]{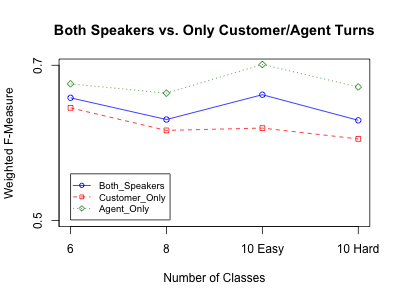}
      \caption{Plot of Both Speaker Turns vs. Only Customer/Agent Turns for Sequential SVM-HMM}
       \label{exp-speaker}
\end{figure}

Our next experiment tests the differences in performance when training and testing our real-time sequential SVM-HMM model using only a single type of speaker's turns (i.e. only {\it Customer} or only {\it Agent} turns). Figure \ref{exp-speaker} shows the relative performance of using only speaker-specific turns, versus our standard results using all turns. 

We observe that using {\it Customer}-only turns gives us lower prediction performance than using both speakers' turns, but that {\it Agent}-only turns actually gives us {\it higher} performance. Since agents are put through training on how to interact with customers (often using templates), agent behavior is significantly more predictable than customer behavior, and it is easier to predict agent turns {\it even without} utilizing any customer turn information (which is more varied, and thus more difficult to predict). 

We again observe a boost in performance at out 10-Class (Easy) set, due to the inclusion of lexically well-defined classes. Notably, we achieve best performance for the 10-Class (Easy) set using only agent turns, where the use of the {\it Apology} and {\it Thanks} classes are both prevalent and predictable.

\subsubsection{Company-Wise vs Company-Independent Evaluation}

In our final experiment, we explore the changes in performance we get by splitting the training and test data based on company domain. We compare this performance with our standard setup for SVM-HMM from our baseline experiments (Figure \ref{exp-baselines}), where our train-test data splitting is company-independent (i.e. all conversations are randomized, and no information is used to differentiate different companies or domains). To recap, our data consists of conversations from four companies from three different industrial domains (one from the telecommunication domain, two from the electronics domain, and one from the insurance domain). We create four different versions of our 6-class real-time sequential SVM-HMM, where we train on the data from three of the companies, and test on the remaining company. We present our findings in Table~\ref{tab:company-wise_evaluation}. 

\begin{table}[h!]
\centering
\small
\caption{Company-Wise vs Company-Independent Evaluation for 6-Class Sequential SVM-HMM}
\label{tab:company-wise_evaluation}
\begin{tabular}{p{3.5cm}|p{1.4cm}|p{2.5cm}}
\bf Experimental Setup & \bf Weighted& \bf Company-Wise \\
			      & \bf F-Measure& \bf Train/Test Fold Size \\\hline
Test-Electronics-1, Train Rest & 0.632 & 493 / 307\\
Test-Electronics-2, Train Rest & 0.599 & 592 / 208\\
Test-Telecom, Train Rest & 0.585 & 604 / 196\\
Test-Insurance, Train Rest & 0.523 & 711 / 89 \\ \hline
Company-Independent & 0.658 & - \\
\end{tabular} 
\end{table}

\begin{figure*}[h!]
\centering
\begin{subfigure}{.33\textwidth}
  \centering
    \includegraphics[width=1\textwidth]{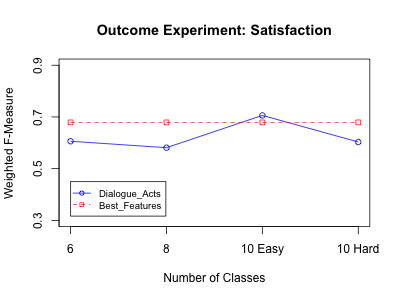}
  \caption{Satisfaction Outcome}
  \label{fig:sfig1}
\end{subfigure}%
\begin{subfigure}{.33\textwidth}
  \centering
    \includegraphics[width=1\textwidth]{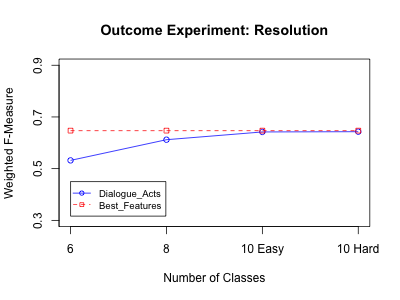}
  \caption{Resolution Outcome}
  \label{fig:sfig2}
\end{subfigure}
\begin{subfigure}{.33\textwidth}
  \centering
    \includegraphics[width=1\textwidth]{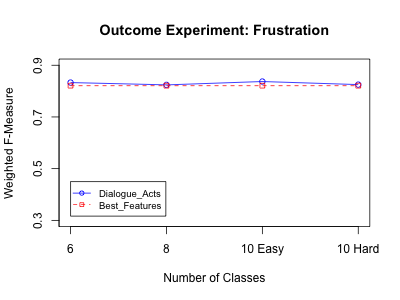}
  \caption{Frustration Outcome}
  \label{fig:sfig3}
\end{subfigure}
\caption{Plot of Dialogue Act Features vs. Best Feature Sets for Satisfaction, Resolution, and Frustration Outcomes}
\label{exp-outcomes}
\end{figure*}

From the table, we see that our real-time model achieves best prediction results when we use one of the electronics companies in the test fold, even though the number of training samples is smallest in these cases. On the other hand, when we assign insurance company in the test fold, our model's prediction performance is comparatively low. Upon further investigation, we find that customer-agent conversations in the telecommunication and electronics domains are more similar than those in the insurance domain. Our findings show that our model is robust to different domains as our test set size increases, and that our more generic, company-independent experiment gives us better performance than any domain-specific experiments. \\

\section{Conversation Outcome Analysis}
\label{outcome-analysis}

Given our observation that {\it Agent} turns are more predictable, and that we achieve best performance in a company-independent setting, we question whether the training that agents receive is actually reliable in terms of resulting in overall "satisfied customers", regardless of company domain. Ultimately, our goal is to discover whether we can use the insight we derive from our predicted dialogue acts to better inform conversational systems aimed at offering customer support. Our next set of experiments aims to show the utility of our real-time dialogue act classification as a method for summarizing semantic intent in a conversation into rules that can be used to guide automated systems. 

\subsection{Classifying Problem Outcomes}
We conduct three supervised classification experiments to better understand full conversation outcome, using the default Linear SVC classifier in Scikit-Learn \cite{scikit-learn} (which gave us our best baseline for the dialogue classification task). Each classification experiments centers around one of three problem outcomes: customer satisfaction, problem resolution, and customer frustration. For each outcome, we remove any conversation that did not receive majority consensus for a label, or received majority vote of "can't tell". Our final conversation sets consist of 216 {\it satisfied} and 500 {\it unsatisfied} customer conversations, 271 {\it resolved} and 425 {\it unresolved} problem conversations, and 534 {\it frustrated} and 229 {\it not frustrated} customer conversations. We retain the inherent imbalance in the data to match the natural distribution observed. The clear excess of consensus of responses that indicate negative outcomes further motivates us to understand what sorts of dialogic patterns results in such outcomes.

We run the experiment for each conversation outcome using 10-fold cross-validation, under each of our four class settings: 6-Class, 8-Class, 10-Class (Easy), and 10-Class (Hard). The first feature set we use is {\it Best\_Features} (from the original dialogue act classification experiments), which we run as a baseline. 

{ Our second feature set is our {\it Dialogue\_Acts} predictions for each turn -- we choose the most probable dialogue act prediction for each turn using our dialogue act classification framework to avoid sparsity. In this way, for each class size $N$, each conversation is converted into a vector of $N$ (up to 10) features that describe the most strongly associated dialogue act from the dialogue act classification experiments for each turn, and the corresponding turn number. For example, a conversation feature vector may look as follows: 
\[
\begin{bmatrix}
{\tt statement\_complaint:turn\_1} \\
{\tt request\_info:turn\_2}\\
{\tt ...} \\
{\tt greeting\_closing:turn\_N} \\
\end{bmatrix}
\]

Thus, our classifier can then learn patterns based on these features (for example, that specific acts appearing at the end of a conversation are strong indicators of customer satisfaction) that allow us to derive rules about successful/unsuccessful interactions.}

Figure \ref{exp-outcomes} shows the results of our binary classification experiments for each outcome. For each experiment, the {\it Best\_Features} set is constant over each class size, while the {\it Dialogue\_Act} features are affected by class size (since the predicted act for each turn will change based on the set of acts available for that class size). Our first observation is that we achieve high performance on the binary classification task, reaching F-measures of 0.70, 0.65, and 0.83 for the satisfaction, resolution, and frustration outcomes, respectively. Also, we observe that the performance of our predicted dialogue act features is comparable to that of the much larger set of best features for each label (almost identical in the case of frustration).

{ In more detail, we note interesting differences comparing the performance of the small set of dialogue act features that "summarize" the large, sparse set of best features for each label, as a form of data-driven feature selection.} For satisfaction, we see that the best feature set outperforms the dialogue acts for each class set except for 10-Class (Easy), where the dialogue acts are more effective. The existence of the very lexically well-defined {\it Social Act Thanking} and {\it Social Act Apology} classes makes the dialogue acts ideal for summarization. In the case of problem resolution, we see that the performance of the dialogue acts approaches that of the best feature set as the number of classes increases, showing that the dialogue features are able to express the full intent of the turns well, even at more difficult class settings. Finally, for the frustration experiment, we observe negligible different between the best features and dialogue act features, and very high classification results overall.

\subsection{Actionable Rules for Automated Customer Support}
\begin{table*}[h!]
\centering
\small
\caption{Most Informative Dialogue Act Features and Derivative Actionable Insights, by Conversation Outcome}
\label{outcome-features}
\begin{tabular}{p{2.5cm}|p{8cm}|p{1.5cm}|p{2cm}|p{1cm}}
\bf Dialogue Act & \bf Example & \bf Position & \bf Outcome & \bf Weight \\ \hline
Offering more help & We can contact you if you like. Please DM us your info. & End & Satisfied & 13.25 \\
& & & Resolved & 8.72 \\ \hline

Thanking & Thanks for your reply! We're happy to hear your issue & End & Satisfied & 8.58 \\
& has been taken care of. & & Resolved & 6.77 \\ \hline

Answer (Other) & You can view our available selection of iPhone 6 & Start & Satisfied & 3.46 \\
& cases here $<$link$>$  :) & & Not Frustrated & 3.5 \\ \hline

Apology & I do apologize for the inconvenience. Please provide me with the cross streets and zip code, so I can look into this for you. & Mid & Satisfied & 3.31 \\ \hline

Suggestion & Oh that's odd! Let's start by power cycling & Start & Resolved & 5.98 \\ 
& your console: $<$link$>$  & & Not Frustrated & 3.52 \\ \hline

Question Yes-No & I'm sorry if you were unable to speak with a manager. & Start & Resolved & 3.30 \\
& Are there any questions I can assist you with? & & Not Frustrated & 3.39 \\ \hline

Statement Info & I'm so sorry you were given conflicting info. The discounted iphone 6 starts at \$199.99 with a new 2-year contract. & Start & Not Frustrated & 4.29 \\ \hline\hline

\rowcolor{Gray}
Suggestion & We do not DM, but you can send the details to& End & Unsatisfied & 5.45 \\
\rowcolor{Gray}
&  twitter@[agent].com - we should be able to look into this. & & Unresolved & 5.98 \\\hline

\rowcolor{Gray}
Request Info & Please DM us your name, state, and policy number so we can & End & Unsatisfied & 4.32 \\
\rowcolor{Gray}
& have someone review your policy for discounts.  & & Unresolved & 5.30 \\\hline

\rowcolor{Gray}
Question Yes-No & Hmm, does the messaging showing match any of those on this page: $<$link$>$? & End & Unsatisfied & 3.19 \\\hline

\rowcolor{Gray}
Complaint & Well what I received last night was not high quality service. & End & Unsatisfied & 3.00 \\
\rowcolor{Gray}
 & Especially for what I pay. You will be hearing from me. & Mid & Frustrated & 8.56 \\ \hline
 
\rowcolor{Gray}
 Apology & I'm sorry to hear about this. Have you checked your spam mail? & Start & Frustrated & 2.92 \\\hline
 
\rowcolor{Gray}
 Statement Info & No, normally you're supposed to receive them within minutes. & End & Frustrated & 5.35 \\\hline
 
\rowcolor{Gray}
 Expressive Neg. & I reinstated online. Just mad was told would not be cancelled, and it happened anyway. & Mid & Unresolved & 2.58 \\ \hline
\end{tabular}
\end{table*}

While these experiments highlight how we can use dialogue act predictions as a means to greatly reduce feature sparsity and  predict conversation outcome, our main aim is to gain good insight from the use of the dialogue acts to inform and automate customer service interactions. We conduct deeper analysis by taking a closer look at the most informative dialogue act features in each experiment.

Table \ref{outcome-features} shows the most informative features and weights for each of our three conversation outcomes. To help guide our analysis, we divide the features into positions based on where they occur in the conversation: {\it start} (turns 1-3), {\it middle} (turns 4-6), and {\it end} (turns 7-10). Desirable outcomes (customers that are satisfied/not frustrated and resolved problems) are shown at the top rows of the table, and undesirable outcomes (unsatisfied/frustrated customers and unresolved problems) are shown at the bottom rows.

Our analysis helps zone in on how the use of certain dialogue acts may be likely to result in different outcomes. The weights we observe vary in the amount of insight provided: for example, offering extra help at the end of a conversation, or thanking the customer yields more satisfied customers, and more resolved problems (with ratios of above 6:1). However, some outcomes are much more subtle: for example, asking yes-no questions early-on in a conversation is highly associated with problem resolution (ratio 3:1), but asking them at the {\it end} of a conversation has as similarly strong association with {\it unsatisfied} customers. Giving elaborate answers that are not a simple affirmative, negative, or response acknowledgement (i.e. Answer (Other)) towards the middle of a conversation leads to satisfied customers that are not frustrated. Likewise, requesting information towards the end of a conversation (implying that more information is still necessary at the termination of the dialogue) leads to unsatisfied and unresolved customers, with ratios of at least 4:1. 

By using the feature weights we derive from using our predicted dialogue acts in our outcome classification experiments, we can thus derive data-driven patterns that offer useful insight into good/bad practices. Our goal is to then use these rules as guidelines, serving as a basis for automated response planning in the customer service domain. For example, these rules can be used to recommend certain dialogue act responses given the position in a conversation, and based previous turns. This information, derived from correlation with conversation outcomes, gives a valuable addition to conversational flow for intelligent agents, and is more useful than canned responses.


\section{Conclusions}
\label{conclusions}
In this paper, we explore how we can analyze dialogic trends in customer service conversations on Twitter to offer insight into good/bad practices with respect to conversation outcomes. We design a novel taxonomy of fine-grained dialogue acts, tailored for the customer service domain, and gather annotations for 800 Twitter conversations. We show that dialogue acts are often semantically overlapping, and conduct multi-label supervised learning experiments to predict multiple appropriate dialogue act labels for each turn in real-time, under varying class sizes. { We show that our sequential SVM-HMM model outperforms all non-sequential baselines, and plan to continue our exploration of other sequential models including Conditional Random Fields (CRF) \cite{crf} and Long Short-Term Memory (LSTM) \cite{lstm}, as well as of dialogue modeling using different Markov Decision Process (MDP) \cite{mdp} models such as the Partially-Observed MDP (POMDP) \cite{pomdp}}. 

We establish that agents are more predictable than customers in terms of the dialogue acts they utilize, and set out to understand whether the conversation strategies agents employ are well-correlated with desirable conversation outcomes. We conduct binary classification experiments to analyze how our predicted dialogue acts can be used to classify conversations as ending in customer satisfaction, customer frustration, and problem resolution. We observe interesting correlations between the dialogue acts agents use and the outcomes, offering insights into good/bad practices that are more useful for creating context-aware automated customer service systems than generating canned response templates.

Future directions for this work revolve around the integration of the insights derived in the design of automated customer service systems. To this end, we aim to improve the taxonomy and annotation design {by consulting domain-experts} and using annotator feedback and agreement information, derive more powerful features for dialogue act prediction, and automate ranking and selection of best-practice rules based on domain requirements for automated customer service system design.

\balance{}

\bibliographystyle{SIGCHI-Reference-Format}
\bibliography{dialogue_acts_v1}
\end{document}